\documentclass[technote]{IEEEtran}
\IEEEoverridecommandlockouts
\usepackage{cite}
\usepackage{amsmath,amssymb,amsfonts}
\usepackage{algorithm}
\usepackage{algpseudocode}
\usepackage{graphicx}
\usepackage{textcomp}
\usepackage{xcolor}
\usepackage{url}
\def\BibTeX{{\rm B\kern-.05em{\sc i\kern-.025em b}\kern-.08em
    T\kern-.1667em\lower.7ex\hbox{E}\kern-.125emX}}
\begin{document}

\title{Re-determinizing Information Set Monte Carlo Tree Search in Hanabi}

\author{\IEEEauthorblockN{James Goodman}
james@janigo.co.uk}

\maketitle

\begin{abstract}
This technical report documents the winner of the Computational Intelligence in Games(CIG) 2018 Hanabi competition. We introduce Re-determinizing IS-MCTS, a novel extension of Information Set Monte Carlo Tree Search (IS-MCTS) \cite{IS-MCTS} that prevents a leakage of hidden information into opponent models that can occur in IS-MCTS, and is particularly severe in Hanabi.
Re-determinizing IS-MCTS scores higher in Hanabi for 2-4 players than previously published work at the time of the competition. 
Given the 40ms competition time limit per move we use a learned evaluation function to estimate leaf node values and avoid full simulations during MCTS.
For the Mixed track competition, in which the identity of the other players is unknown, a simple Bayesian opponent model is used that is updated as each game proceeds. 
\end{abstract}

\begin{IEEEkeywords}
Hanabi, Monte Carlo Tree Search, Opponent Modelling, Information Sets
\end{IEEEkeywords}

\section{Introduction}
{\let\thefootnote\relax\footnote{This report complements a paper accepted for the IEEE Conference on Games (CoG), London, United Kingdom, 2019; and provides significantly more implementation details}}
Hanabi \cite{bgghanabi} is a co-operative game for 2-5 players that has attracted some attention in games research due to the role of hidden information, a restricted communication channel and need to model one's fellow players, most recently \cite{DeepMindHanabi}. The Hanabi-agent competition at the Computational Intelligence in Games (CIG) 2018 conference had two tracks. The Mirror track has all players in a game using the same agent. This makes modelling of other agents relatively straightforward, but requires a strategy for communicating hidden information to other players. The Mixed track has a random set of unknown different agents playing each game, including other competition entrants, and to do well it is necessary to model the strategies being used by the other players.

Information Set Monte Carlo Tree Search (IS-MCTS) \cite{IS-MCTS} performs very poorly in Hanabi, regardless of time budget. We trace this pathology to the specific form of hidden information in Hanabi, and address it with a new variant of IS-MCTS that we term Re-determinizing IS-MCTS (RIS-MCTS) that avoids leakage of hidden information known to the acting player into the modelling of other players in the tree search. This is done by re-determinizing hidden information from the perspective of the acting player at each node in the tree search (to be distinguished from the active player in the \emph{game}, who is always the root player in the tree).

To achieve a good standard of play further amendments to vanilla MCTS are needed to reduce the searched action space via rule heuristics, and iteratively learn a simulation policy from the result of off-line RIS-MCTS games. To meet the competition constraint of 40ms per decision, the final competition entry avoids full simulations during search, and evaluates the expanded node(s) in the tree using the learned function.

As well as describing the winning entry, this report offers two specific contributions. Firstly the Re-determinizing IS-MCTS algorithm, as the information leakage problem identified in Hanabi is likely to be present in other environments. We note that RIS-MCTS has theoretical flaws that we discuss in Section \ref{discussion}, but these do not prevent state-of-the-art performance in Hanabi.
Secondly we introduce a method of training state evaluation functions from offline MCTS games that uses more of the data in an MCTS tree than just the root node.

\begin{figure}[!t] 
	\centering
	\label{hanabiGame}
	\includegraphics[width=3in]{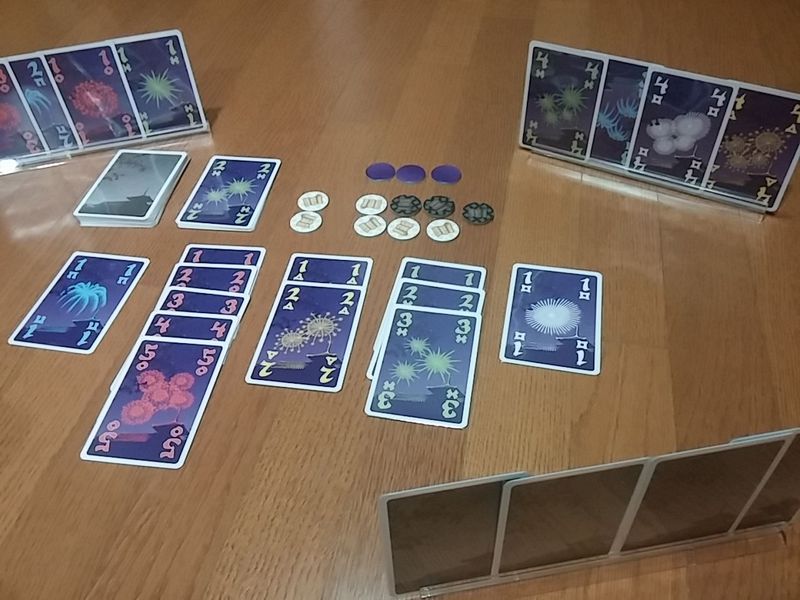}
	\caption{A game of Hanabi in progress. The player at the camera's perspective can see the other players' cards, but not their own. The current score in the game is 12, from the sum of the top cards in each suit in the tableau.}
\end{figure}

We first cover background and key previous work in both Hanabi and MCTS in Sections \ref{background} and \ref{previousWork}. We then describe the detailed implementations for the winning Mirror Track entry in Sections \ref{implementation} (the RIS-MCTS algorithm), \ref{furtherImprovements} (additional game-specific improvements needed), and \ref{Mirror} (fitting the computation into 40ms). Section \ref{Mixed} then describes the changes needed in opponent modelling for the Mixed Track. Section \ref{discussion} analyses in more detail why RIS-MCTS works well in Hanabi despite its identified flaws, and areas for further work in extending RIS-MCTS to more general imperfect information games. Section \ref{conclusion} concludes, and also discusses further work specific to play of Hanabi. All code for the entries to both tracks is available at \url{https://github.com/hopshackle/fireworks}.

\section{Background} \label{background}

\subsection{Hanabi}
Hanabi has a deck of 50 cards. There are five suits (red, blue, green, white, yellow), and five possible face values (one through five). Each suit has 10 cards - one `five' card, three `one' cards, and two of each other face value. The goal is to play out cards to a shared tableau  so that each suit is played out in order, starting with the `one' card through to the `five'.
The final score is the sum of the top face-value cards in each suit, for a maximum score of 25.
The game is co-operative with all players trying to obtain the highest communal score.
Each player in Hanabi is dealt a hand of 4 or 5 cards (depending on the number of players), but holds the cards to face the other players without looking at them. Each player therefore knows exactly which cards are held by their fellow players, but does not know what their own cards are. 
The remaining cards after the initial deal form a face-down deck.
A game of Hanabi in progress is shown in Figure \ref{hanabiGame}.

Play proceeds clockwise round the table, and on their turn a player may do one of three things:
\begin{itemize} 
	\item{Play a card from their hand to the tableau;}
	\item{Discard a card from their hand;}
	\item{Give a hint to another player by touching all of their cards that meet a stated criterion; either all cards of a stated colour, or all cards with a stated face value.}
\end{itemize}
If a card played to the tableau is not in sequence, then one life is lost. After three lives are lost the game is over. 
When a card is played or discarded, the player draws a new unseen card from the deck. Once the last card is drawn from the deck, each player has one last turn and then the game is also over.
There is a communal limit on the number of hints, with 8 hint tokens available. When a hint is given, one of these is used up; every time a card is discarded or a `five' card successfully played, then a hint token is regenerated. If no hint token is available, then a player must either Play or Discard a card.
A hint must refer to a card colour or face value that the player possesses - e.g. it is not possible to tell a player that they have no red cards in their hand. No communication about hand contents is permitted apart from the hints.

Baffier et al. \cite{baffier} show formally that Hanabi is NP-complete, and not all games can achieve a perfect score of 25 even if  players could see all the cards and the deck ordering.
The challenge in the game is to make efficient use of the hint tokens, and model the thinking of the other players, both to predict their future actions and infer from their hints what one's own cards are.

\subsection{MCTS}
Monte Carlo Tree Search (MCTS) is surveyed in detail in \cite{Browne}, and has been used successfully in many games, such as Go \cite{AlphaZero}, Lines of Action \cite{loa}, Settlers of Catan \cite{catan}, Magic: The Gathering \cite{mtg}, Hearthstone \cite{hearthstoneMCTS} \cite{hearthstoneZero} and Scotland Yard \cite{ScotlandYard}.
It is an \emph{anytime} algorithm that uses an available time budget to stochastically search the forward game tree from the current state. On each iteration four steps are followed:
\begin{enumerate}
\item{Selection. Select an action to take from the current state. If all actions have been selected at least once then the `best' one is picked (usually, as in this work, using the Upper Confidence equation (\ref{UCT}) \cite{Szepesvari}), and this step is repeated down the tree of game states until a state is reached with previously untried actions.
\begin{equation}\label{UCT}
J(a) = V(a) + C \sqrt{\frac{\log{N}}{n(a)}}
\end{equation}
In (\ref{UCT}), the action with largest $J(a)$ is selected at each tree node. $N$ is the total number of visits (iterations through) the node; $n(a)$ is the number of those visits that then took action $a$; $V(a)$ is the  mean score for all visits to the node that took action $a$; $C$ is a parameter that controls the trade-off between exploitation using the empirical score $V(a)$, and exploration choosing actions with few visits so far.
}
\item{Expansion. Pick one of the untried actions at random, and expand this, creating a new node in the game tree.}
\item{Simulation. From the expanded node, simulate a complete game to obtain a final score, for example by taking moves at random.
During the simulation step, domain knowledge is commonly embodied in a simulation policy rather than using random moves, e.g. \cite{GellySilver2007} \cite{hearthstoneMCTS} \cite{catan}.
}
\item{Back-propagation. Back-propagate this final score up the game tree. Each node records the mean score of all iterations that take a given action from that node as $V(a)$ in (\ref{UCT}), that will affect future Selection steps.}
\end{enumerate}

Once the available time budget has been used up the action at the root node with the highest score is then picked and executed in the actual game environment. There are many variants for each of the four key steps above. We reference the most relevant previous work when detailing the implementation in Section \ref{implementation}, and see \cite{Browne} for a comprehensive survey.

\section{Previous Work} \label{previousWork}
\subsection{MCTS with hidden information}\label{ISMCTS}

MCTS requires a forward model of the game, so that when an action is selected at a node, the game state can be rolled forward to a hypothetical `what-if' game state at the child node. In a game with hidden information we cannot roll forward to a single game state, as this depends on unknown information. For example in Hanabi, if we play our (unknown) second card from hand, the `what if' rolled forward state will vary based on what the actual card turns out to be. 
One approach is to sample a possible set of hidden information as a game `determinization', and proceed as if this information is known. Determinizations are sampled from the player's information set (the set of all game states that are indistinguishable to the player) and MCTS applied independently to each one as a perfect information game. A final action is chosen by averaging across the statistics at the root nodes for each determinized game. This Perfect Information Monte Carlo (PIMC) has been used in Klondike Solitaire \cite{Klondike} and Bridge \cite{gib}.

The major problems with determinization (converting to a single perfect information game) are \emph{strategy fusion} and \emph{non-locality} \cite{FrankBasin}. Strategy fusion occurs when an agent makes different decisions from the same information set due to implicit or explicit use of hidden information. 
For example, if use MCTS to solve several determinizations of a game (as in PIMC), then in each case MCTS will determine the single best action to take, and these will vary with the determinization despite the player being in the same information set. We \emph{hope} that averaging over these mutually incompatible suggestions will give a good answer, but the best move might in fact be one that gathers more information about the actual hand the opponent has so we can make a more informed decision later on; PIMC will never find this.
Non-locality arises because the likely values of hidden information will depend on historic moves in the game - this is not a problem in perfect information environments, which are sub-game perfect and can be decomposed into the solution of their sub-games \cite{Bitan}. For example, in Whist it is unusual to lead away from an Ace; so if an opponent leads a low heart, we would expect their hand (the hidden information) to probably not include the Ace of Hearts.

Monte Carlo Search in Partially Observable Markov Decision Processes (POMDPs) with a single-player \cite{SilverVeness} addresses strategy fusion with a particle filter of possible determinizations constructed at the root, which is sampled from the current information set. Each iteration uses a single determinization from this pool, and maintains a set of possible determinizations at each node in the tree. If a particular card in hand has a 20\% chance of being playable, then it will be playable in 20\% of the simulations. This also addresses non-locality as after each action and transition in the real game the set of determinizations at the child node is used to seed the particle filter for the next action, defining a non-uniform distribution for the new information set.

Information Set MCTS \cite{IS-MCTS} is introduced in two forms. Single Observer (SO-ISMCTS) and Multiple Observer (MO-ISMCTS). Both, like \cite{SilverVeness}, sample a different possible determinization at the root of the tree for each MCTS iteration, and maintain a node for each information set from the perspective of the current player.

In SO-ISMCTS opponents make random moves consistent with the current determinization. MO-ISMCTS improves opponent modelling by using IS-MCTS for their moves as well and constructs a tree for each player.
With MO-ISMCTS the decisions made by other players are still made with respect to determinizations from the root player's perspective, and non-locality is not addressed.
Cowling et al. \cite{Resistance} extend MO-ISMCTS to the bluffing game \emph{The Resistance} with improved opponent modelling. This algorithm (MT-ISMCTS) requires one tree to be maintained per opponent for each information set that they could be in, and these are updated using determinizations that the acting player knows to be incorrect, but which \emph{could} be correct from the perspective of the other players. This produces a much better opponent model. In Resistance there are never more than six possible information sets for a given player. In Hanabi the number of information sets that another player could be in is defined by what cards the current player is holding, which at the start of the game is $\sim10^5$ with 4 unknown cards, making this approach of one tree per information set less tractable.

Semi-Determinized MCTS (SDMCTS) \cite{Bitan} extends IS-MCTS to include inference as to which individual states in an Information Set are more likely and address non-locality. This determinizes just the opponent's previous move, and runs IS-MCTS separately for each possibility to calculate an estimated score. An opponent model pre-learned from human play traces is then used to predict a distribution over the opponent's previous move and calculate an expected best response.
Nijssen et al. \cite{ScotlandYard} bias the determinizations in MCTS using statistics from self-play games to similarly address non-locality



\subsection{Hanabi}\label{HanabiHistory}
The first work on playing Hanabi is from 2015, when Ozawa et al. \cite{Ozawa} investigate 2-player Hanabi strategies, and compare some heuristic rules with an approach that models the other player's likely next action assuming they are using the same strategy. This improves the score to 15.9 from 14.5 with heuristics only.

Van den Bergh et al. 2016 \cite{vdb} further investigate heuristic rules in 3-player Hanabi, and search over a set of 48 combinations to find the best-scoring. The winner plays a card when 60\% confident it is playable; else discards a card when 100\% confident it can be discarded; else hints about a card that is immediately playable; else hints about the most cards possible; else discards the card most likely to be discardable. They also try MCTS with 500 iterations, but find that the average performance here is lower (14.5 vs 15.4) than for the simple heuristic rule. 
While not formally using IS-MCTS the authors note that with MCTS care must be taken to avoid leaking hidden information to later players. For this reason they re-shuffle the hand of the acting player at each \emph{move} in tree selection to one that is compatible with the hints given so far, while IS-MCTS only shuffles once per tree iteration to determinize the hand of the root player. They do not use complete game simulations, but record the number of points gained over the next 5 moves, assuming infinite lives with a random simulation policy.

Cox et al. 2015 \cite{Cox} have a very different approach that uses hat-guessing to pass on information to other players. This assigns a value (the `hat') to each hand that instructs the player which card is playable or discardable to give a number between 0 and 2C - 1 (where C is the number of cards in a hand). The sum of this over all hands, modulus the number of players, is then used via a pre-agreed lookup table to define a hint to give in terms of player, colour and face value. After one round of Hint actions, each player can then precisely calculate (by working out their own `hat') which card in their hand to play or discard. However this only works with 5-players.
Bouzy 2017 \cite{Bouzy} extends this hat-guessing to different number of players and obtains excellent results, but requires one  Hanabi rule to be dropped - the rule that forbids a hint to touch no cards\footnote[1]{This rule was ambiguous in the first versions of Hanabi, and the designer clarified in 2010 that he had not intended to explicitly forbid a `no touch' hint \cite{BauzaBGG}. However, from at least 2014 the official rulebook has explicitly forbidden these \cite{hanabi2014}, and this official ruleset is the one applicable to the competition.}. Without this relaxation of the rules (and increase in the available information bandwidth) it can be impossible for a required hint to be made, and the hat-guessing approach fails. 

As well as the hat-guessing approach, \cite{Bouzy} uses 1-ply expectimax search with 1000 different determinizations of the current state to approximate the hidden information distribution, and simulating all of these for each possible action using a variety of heuristic simulation policies. The paper finds that the best result uses a `confidence' simulation policy that assumes a card hinted by a previous player is playable.

Walton-Rivers et al. \cite{Walton-Rivers} find that pure MO-ISMCTS that uses tree search to take actions for all the players gives poor results, and look at explicit modelling of the other agents in Hanabi. They use an IS-MCTS agent that models the other players explicitly using heuristic rules. This means that nodes in the tree are only from the perspective of the acting player, with the actions of the other players subsumed into the environment.

Eger et al. 2017 \cite{Eger} look at Hanabi from the perspective of designing an AI agent to play with humans. They find that human players prefer playing with an agent that gives hints that are immediately relevant to play, such as hinting an immediately playable or discardable card, rather than those that maximise information in the longer term.

Canaan et al. 2018 \cite{Canaan} use genetic algorithms to evolve a heuristic constructed from an ordering of rules inspired by those of \cite{Walton-Rivers}, and using the same framework. They evolve rules that surpass previous work (excluding the hat-guessing algorithm of \cite{Cox}), and came second in both tracks of the CIG 2018 Hanabi competition.

Most recently Bard et al. 2019 \cite{DeepMindHanabi}, Foerster et al. 2018 \cite{BAD} use Deep Reinforcement Learning to learn conventions
and achieve state of the art performance in 2-player Hanabi. They document hand-coded bots from outside the academic literature that are state of the art with 3 or more players (see Table \ref{soa}), and use a strictly harder Hanabi variant in which losing all lives gives zero points and not the current tableau score.

\section{Re-determinizing Information Set MCTS (RIS-MCTS)} \label{implementation}
\subsection{Why MO-ISMCTS fails in Hanabi}\label{ISMCTSfailure}
MO-ISMCTS determinizes at the root node for each iteration, randomising the hand of the current player as the other players' hands are fully known. 

This avoids strategy fusion occurring for the current (root) player, but not when it comes to the actions of the other players. For example if the current player is A, and B has a playable Red Two (R2) in their hand, then regardless of what player A chooses to do, when we reach player B in the search tree a positive reward will always be received for playing that card, despite the fact that player B cannot possibly know this. This renders any hints that player A gives meaningless, as they have no impact either on the available actions or the action consequences for the player receiving the hint.
This strategy fusion problem in the opponent model is inherent in MO-ISMCTS given the use of a single determinization at the root of all the player trees, and information leaks out to inform the moves made by opponents - we model their behaviour as if they know what we know (which they don't).
The node for player B (representing an Information Set) should be seen as reachable from a number of very different hypothetical game-states, in most of which the given card is \emph{not} R2, or even playable. MO-ISMCTS does not account for this and suggests different actions in different games from what is the same information set for player B: strategy fusion.

\subsection{Single vs Multiple Trees in ISMCTS}
Walton-Rivers et al. 2017 \cite{Walton-Rivers} use MO-ISMCTS to construct an Hanabi agent with a single tree for all players, keeping track of which player is acting at each node. 
RIS-MCTS also uses a single tree for all players. This ostensibly differs from the one tree per player specified in \cite{IS-MCTS}, but is equivalent as long as we have fully observable moves. 

To see this consider what happens between each of our moves. All other players make a move, and because in Hanabi they are fully observable we can be in only one Information Set (defined by the moves taken and the cards drawn) when it gets back to our turn\footnote[2]{We ignore the identity of any card drawn during our turn, which formally does increase the hidden information of other players.}. 
Only an increase in hidden information possessed by a player can increase the number of different Information Sets they are in.
Crucially this also holds for each of the other players too, as none of them gain additional hidden information while making their own moves. Hence, regardless of moves made, they too must be in a unique Information Set on our next turn. This means we can interleave all of the players actions into a single tree, as each move leads to to a single node (information set).

In the case of partially observable moves, which IS-MCTS is designed to support, this does not hold. In a partially observable move taken by another player, they have gained private information, such as the specific card they have picked up. This means they may be in a number of different information sets (nodes) that the original player cannot distinguish between, and we cannot merge the trees easily.

\subsection{The RIS-MCTS algorithm}

\begin{algorithm}
	\caption{RIS-MCTS Algorithm outline. The changes to MO-ISMCTS are in the functions {\sc EnterNode}, called on entering each node during Tree Search, and {\sc ExitNode} called on Exiting each node. {\sc Redeterminize} shuffles the player's hand and deck in line with their current information set.}\label{detAlgo}
	\centering
	\begin{algorithmic}[1]
		\Function{RIS-MCTS}{$root$}
		\While{timeAvailable}
		\State $root.hand \gets \Call{redeterminize}{root}$
		\State $rootNode \gets emptyNode$
		\State $node \gets rootNode$
		\State $player \gets root$
		\While {$node \text{ fully expanded}$}
		\State $newNode \gets \Call{SelectUCT}{node}$
		\State $\Call{ExitNode}{player, rootPlayer}$
		\State $player \gets newNode.player$
		\State $node \gets newNode$
		\State $\Call{EnterNode}{player, rootPlayer}$
		\EndWhile
		\State $node \gets \Call{Expand}{node}$
		\State $score \gets \Call{Simulation}{node}$
		\State $\Call{BackPropagate}{score, node}$
		\EndWhile 
		\State \Return $rootNode.bestAction$
		\EndFunction
		\\
		\Procedure{EnterNode}{player, root}
		\If {$player \ne root$}
		\State $savedHand \gets player.hand$
		\State $player.hand \gets \Call{redeterminize}{player}$
		\EndIf
		\EndProcedure
		\\
		\Procedure{ExitNode}{player, root}
		\If {$player \ne root$}
		\State $player.hand \gets savedHand$ 
		\State $\Call{RemoveIncompatibleCards}{player.hand}$
		\State $\Call{DeterminizeEmptySlots}{player.hand}$
		\EndIf
		\EndProcedure
	\end{algorithmic}
\end{algorithm}

One approach with multiple trees is to determinize the state for each player independently, and play down each tree using this determinization. Since all such determinizations cannot be mutually compatible (unless they are identical), we also need to account for what happens if a player makes a move that is illegal from the perspective of another.

MT-ISMCTS \cite{Resistance} addresses this by having a separate tree for each possible information set each other player could be in, but this is not feasible in Hanabi.
Van den Bergh et al. \cite{vdb} do re-determinize the hand of the active player at each node in the tree, and skip over this illegality problem by assuming infinite lives and restricting any rollout to just 5 moves. This means that losing lives, or even ending the game, is no longer penalised in a simulation. This is not ideal.

Our solution to this problem, which we term Re-determinizing IS-MCTS (RIS-MCTS) is to re-determinize the game-state at every node in a similar fashion to the re-shuffle of \cite{vdb}. Specifically, when any player other than the root player has to make a decision in the tree, we first re-determinize their hidden information (in Hanabi, randomize their hand to be any valid set of cards they could have given the information they have received). If they choose to play a given card, continuing the above example, then the card played in-game is the randomly sampled one, and \emph{not} the card that the other players know they actually have. This resolves the information leakage problem - if the sampled card is not in fact playable, then a life is lost, and it is now worthwhile for player A to hint that the card is a `two'.
This does not avoid the illegality problem and determinizations may be incompatible with the root player's information set. Say that player B played a card in the above scenario, and this was sampled to be Yellow Five (Y5), then the Y5 card is now publicly in the discard pile, and a new card is drawn into the slot. This directly contradicts the root node information that R2 was in the slot, and possibly even that Y5 is elsewhere in player B's hand. Regardless of this, we keep Y5 as the card discarded, and from the perspective of the \emph{other} players this iteration has effectively moved into a different game to the one we started in. We continue playing down the tree in this new game with the other players moved into new information sets.

This makes sense if we consider the node for player B to be a node in many hypothetical game trees in the same Information Set from player B's perspective, only one of which is the game currently being played. What we have done is sampled one of these many games (that player A knows to be false), and determinized to that, even if the game-state is now strictly inconsistent with the information sets of the other players in the original game.
When we exit a node we restore the player's hand and game state to the known values in so far as this is possible. This will always be possible if they chose a Hint action, but not if they chose to Play or Discard a card that was different in their re-determinized hand. The card drawn after playing or discarding is drawn after restoring a hand, so that it is always compatible, since the drawing player has no information about it.
Apart from this re-determinization at each node in the tree, the algorithm is single-tree MO-ISMCTS as used in \cite{Walton-Rivers}.
Algorithm \ref{detAlgo} highlights the changes made to the outline of MO-ISMCTS.

The possible Hanabi score range 0-25 is standardized to 0-1, and a $C$-value of $0.1$ is used (equivalent to 2.5 points pre-standardization), as this gave the best results in initial experiments with $C \in \{0.03, 0.1, 0.3, 1, 3\}$. The same value of $C$ was found to be optimal for both RIS-MCTS and MO-ISMCTS.
All experiments were run on a single Google cloud n1 virtual CPU.

The IS-MCTS implementation here and in \cite{Walton-Rivers} uses an Open Loop approach \cite{openloop}, and only differentiates information sets within the tree based on the actions taken by the players. The card drawn from the deck does not change the information set. This is formally incorrect, but would lead to a much larger branching factor, as after each Play or Discard action up to 25 different cards could be drawn, generating 25 distinct IS. These would only differ by one card and share all other information. We treat these as a single node in the tree, and share statistics.  For any given iteration down the tree this will mean that some actions are not possible, and these are ignored - this is already covered in the core IS-MCTS algorithm for partially observable moves \cite{IS-MCTS}.

\begin{table*}
	\caption{MO-ISMCTS and RIS-MCTS in 4-player Hanabi with random and Van Den Bergh (the best performing policy in \cite{vdb}) simulation policies under different time budgets. All players use the same agent. Standard error shown as $\pm$ and 500 random games were run for each setting.}
	\label{MCTSResults}
	\centering
\begin{tabular}{|l|l|r|r|r|r|}
	\hline
	Algorithm & Simulation Policy & 100ms & 300ms & 1000ms & 3000ms \\
	\hline
	MO-ISMCTS & Random & $3.87\pm 0.09$ & $3.94\pm 0.10$ & $3.88\pm 0.10$ & $3.94 \pm 0.10$ \\
	MO-ISMCTS & Van Den Bergh & $11.40\pm 0.13$ & $14.53\pm 0.09$ & $16.79 \pm 0.08$ & $18.36 \pm 0.08$ \\
	\hline
	RIS-MCTS & Random & $4.28 \pm 0.15$ & $4.21 \pm 0.14$ & $4.61 \pm 0.15$ & $5.21 \pm 0.14$ \\
	RIS-MCTS & Van Den Bergh & $11.90 \pm 0.12$ & $14.85 \pm 0.10$ & $16.85 \pm 0.08$ & $18.48 \pm 0.07$ \\
	\hline
\end{tabular}
\end{table*}
 

Table \ref{MCTSResults} shows the improvement using a purely random simulation policy. Vanilla MO-ISMCTS performs very poorly and fails to give any improvement with additional search time - scoring 3.9 for any time budget between 100ms and 3000ms. Any hint given \emph{cannot} affect the results of another player's action, as these use the known values of their cards independently of any hints given.
If we use a heuristic simulation policy instead of a random one, hints will still not directly affect play in the selection phase - but hints given in the tree search \emph{can} now affect the results of the simulation, if the heuristic policy makes use of them.  This explains why using the Van Den Bergh simulation policy is much better and does improve with time. However the raw Van Den Bergh policy achieves a raw score of $17.20\pm 0.08$ by itself with $<1$ms per move, and MO-ISMCTS with 1000ms time budget is still significantly worse than this. 

With RIS-MCTS a random simulation policy does increase its performance as the time budget increases, from 4.3 at 100ms to 5.2 at 3000ms. However, only the tree selection phase can make use of hints, and vanilla RIS-MCTS is still a poor player. Making use of hints in the simulation policy seems much more important, and while RIS-MCTS with a heuristic simulation policy is significantly better with less than 1 second of budget, beyond that it only reaches the same level of performance as MO-ISMCTS with the same simulation policy.

\section{Further Improvements}\label{furtherImprovements}
We now consider further improvements specific to Hanabi that are required to get state-of-the-art performance from RIS-MCTS. We first restrict the branching factor of the tree in \ref{Rules}, and then apply a convention that commonly emerges in human play of the game in \ref{Convention}.
\subsection{Action space restriction with Rules} \label{Rules}

One reason for the poor base performance of RIS-MCTS is the large branching factor in the tree. In a four-player game, where each player has 4 cards we have 4 Discard actions, 4 Play actions, and up to 24 Hint actions (one per other player and card, doubled as we can point to colour or face value), for a maximum branching factor of 32. This means that the search tree does not extend very deeply. RIS-MCTS reaches a decision depth (the mean depth of the deepest node) in the tree of 3.2 at 100ms, increasing to 5.2 at 3000ms (with a simulation policy, the mean depths are 2.5 and 3.8 due to the computational overhead of the simulation policy). This means it is only exploring even tentatively its first move (depth 1), and the immediate moves of up to the next three players.

A human player does not consider all possible moves. They will rarely play a card about which they have no information, and will only consider a small subset of all possible hints, for example cards which are currently playable or discardable and have not yet been hinted. 
We seek to reduce the branching factor by only considering such `sensible' options.
To construct a `sensible' subset of actions, we use the Rules implemented in the Hanabi competition framework (see \cite{Walton-Rivers} for further details). The Rules we use are listed in Table \ref{RuleList}, and in all cases we break ties for a Hint in favour of the next player to play. Each rule defines 0 (if invalid) or 1 action, and these define the possibilities within tree search - no other possible actions are investigated.

\begin{table}
	\caption{Rules used to limit the branching factor in Hanabi}
	\label{RuleList}
	\centering
	\begin{tabular}{|l|p{5cm}|}
		\hline
		Rule & Description \\
		\hline
		TellMostInformation & Tell that provides most information (not previously given) to any other player\\
		TellAnyoneAboutUseful & Tell another player about a playable card \\
		TellDispensable & Tell another player about a card that can be discarded \\
		CompleteTellUseful & Tells another player full information about a playable card (if they already know it is RED, then this will provide missing information that it is a 2) \\
		CompleteTellDispensable & Tell another player full information about a discardable card \\
		CompleteTellUnplayable & Tell another player full information about an unplayable (but not discardable) card \\
		PlayProbablySafe & Play a card if we are at least 70\% confident it is playable\\
		PlayProbablySafeLate & Play a card is we are at least 40\% confident it is playable, and we have 5 or fewer cards left in the deck\\
		DiscardProbablyUseless & Discard the card that the player is most confident is discardable \\
		\hline
		\end{tabular}
\end{table}

\begin{table*}
	\caption{Results for 4-player Hanabi games using RIS-MCTS with Rule-constrained tree options. $\pm$ indicates standard error, and 500 random games were run per setting.  `+ C' indicated use of the `playable now' convention.}
	\label{MCTSRuleResults}
	\centering
	\begin{tabular}{|l|l|r|r|r|r|}
		\hline
		Algorithm & Simulation Policy & 100ms & 300ms & 1000ms & 3000ms \\
		\hline
		MO-ISMCTS & Random & $9.49 \pm 0.11$ & $10.09 \pm 0.11$ & $10.54 \pm 0.10$ & $10.67 \pm 0.10$ \\
		MO-ISMCTS & Van Den Bergh & $7.12 \pm 0.09$ & $7.85 \pm 0.09 $ & $8.30 \pm 0.09$ & $8.97 \pm 0.10$ \\
		\hline
		RIS-MCTS & Random & $17.43 \pm 0.10$ & $17.93 \pm 0.08$ & $18.06 \pm 0.08$ & $18.14 \pm 0.07$ \\
		RIS-MCTS & Van Den Bergh & $17.41 \pm 0.07$ & $18.31 \pm 0.07$ & $19.41 \pm 0.06$ & $19.84 \pm 0.06$ \\
		\hline
		RIS-MCTS + C & Random & $19.40 \pm 0.07$ & $19.67 \pm 0.08$ & $19.84 \pm 0.07$ & $19.76 \pm 0.07$ \\
		RIS-MCTS + C & Van Den Bergh & $17.86 \pm 0.07$ & $19.11 \pm 0.06$ & $20.20 \pm 0.06$ & $20.81 \pm 0.06$ \\
		\hline
	\end{tabular}
\end{table*}

This cuts the maximum branching factor (in a 4-player game) from 32 to 9. As Table \ref{MCTSRuleResults} shows, this dramatically improves the playing strength of the agent, and even with only 100ms of thinking time and a random rollout policy, we outclass the Van Den Bergh heuristic.

For the rules that require a probability of a card being playable or discardable, this is calculated by considering the possible cards it could be, taking into account all visible cards as well as the hints they have been given about the card from other players. There is no attempt here at indirect inference based on hints not given. For example, if as the first move player 1 hints to player 3 that they have two Red cards, player 2 does not infer (as a human player might) that none of their cards are 1s.

The results of an ablative study on the relative impact of restricting the action space via rule heuristics, and the RIS-MCTS changes are shown in Table \ref{MCTSRuleResults}. When we use rule-based action restriction with MO-ISMCTS, hints made within the tree can now affect later actions, as the rules will use them to determine which actions are available later in the tree (the random simulation policy is not rule constrained).
This helps the random simulation policy, but the Van Den Bergh policy gives much worse results than with the full action space. RIS-MCTS now provides a much bigger benefit of 8-11 points over MO-ISMCTS, and beats the Van Den Bergh benchmark (of 17.2) even with 100ms budget. The decision depth has increased to 6.7 to 11.0 with a random simulation policy, and 4.9 to 7.7 with the Van Den Bergh policy. 

Other methods have been used to restrict the explored action space. Progressive un-pruning or widening \cite{progressiveWidening} uses a domain-specific heuristic to initially reduce the actions in the selection step so that only the best ones (according to the heuristic) are available, and actions are added to this as the number of visits to the node increases. This focuses the search on the actions highlighted by the heuristic, but investigates all actions as the time-budget increases. Here we use the rules to provide the heuristic, and we do no un-pruning.

\subsection{The Power of Convention} \label{Convention}
With its restricted communication channel to pass on hidden information to team-members, Hanabi has some similarities to Bridge with a single team instead of two opposed teams. As in Bridge, when a group of humans play Hanabi conventions frequently form. One common example is the convention that a hint will be given to highlight an immediately playable card. If player A tells player B that they have a single blue card, then in the absence of other information player B will frequently play this card, despite formally having no information about its face value. A convention overloads the hint with additional information that `this card is playable now'. 
This also means that other hints - for example hinting that a card is red, when it is not currently playable - cannot be given if the convention is in use. A convention is useful if the hints that it prevents would rarely be used in practise, and hence gives a net increase in the effective communication capacity.

This `playable now' convention is used by \cite{Eger} in designing an `intentional' AI for Hanabi, that plays nicely with human agents; in the `confidence' method of Bouzy 2017 \cite{Bouzy} and is one of the rules that Canaan et al. \cite{Canaan} find helpful in their evolutionary search. 
We adopt a simple convention that if a player gives a hint about a \emph{single} card to the \emph{next} player, then this card is immediately playable (unless this is impossible with information the player has). This does not affect the rules in Table \ref{RuleList}, but affects how the player calculates the probability that a card is playable or discardable. For example, if player A hints to player B about a single Red card they hold, and player B has no other information about this card, then player B will know this is 100\% playable - unless all the playable red cards are visible elsewhere, in which case they know it is 100\% discardable. 
We modify the rules to avoid giving `illegal' hints that would be false under the convention. Implementing this convention increases performance of our RIS-MCTS agents considerably as shown in Table \ref{MCTSRuleResults} as `RIS-MCTS + C', and increases the average score by up to 2 points.
We tried some experiments with a similar convention for discardable cards, but this did not improve performance.

Table \ref{soa} compares the results of previous published work on Hanabi with the performance of our `RIS-MCTS + C' algorithm.
The results for Van den Bergh \cite{vdb} are obtained using the re-implementation of their rules in the Hanabi CIG 2018 framework (and see \cite{Walton-Rivers}), as the original paper only reports for 3-player games with a score of 15.4. The scores for `Bouzy 2017' use that paper's `Confidence' heuristic, and are not perfectly comparable as they permit illegal Hints, as discussed in \ref{HanabiHistory}. Bouzy's Hat-guessing algorithm makes extensive use of these illegal Hints, and we hence exclude it.
The use of 1s and 10s budgets for the RIS-MCTS algorithms is motivated by 1 second being an appropriate time for play with humans, and 10s being the effective budget per move that Bouzy's Expectimax Search algorithm uses in \cite{Bouzy}.

\begin{table}
	\caption{Results for 2 to 5 player Hanabi games. Best published results in bold. Standard error on RIS-MCTS is $\pm 0.2$ for 1s (200 games) and $\pm0.3$ for 10s (100 games). `+ C' uses the `playable now' convention; `+ vdb' uses Van Den Bergh simulation.}
	\label{soa}
	\centering
	\begin{tabular}{|l|l|c|c|c|c|}
		\hline
		Algorithm & 2-P & 3-P & 4-P & 5-P \\
		\hline
		Ozawa et al. 2015  \cite{Ozawa} & 15.9  & - & - & - \\
		Van den Bergh 2015 \cite{vdb} & 13.8 & 17.7 & 17.2 & 16.3 \\
		Cox et al. 2015\cite{Cox}  & - & - & - & 24.7 \\
		Bouzy 2017 \cite{Bouzy} & 15.9 & 17.9 & 19.7 & 19.2 \\
		\emph{Bouzy Expectimax 2017} \cite{Bouzy} & 19.0 & 20.4 & 21.1 & 20.4 \\
		Canaan et al. 2018 \cite{Canaan}  & 20.1 & 19.6 & 19.4 & 18.3 \\
		Foerster et al. 2018 \cite{BAD} & \textbf{23.9} & - & - & - \\
		Bard et al. 2019 \cite{DeepMindHanabi} & 22.7 & 20.2 & 21.6 & 16.8 \\
		\emph{RIS-MCTS (1s)}& 17.7 & 18.6 & 18.1 & 17.0 \\
		\emph{RIS-MCTS (10s)}& 17.9 & 18.9 & 18.2 & 17.1 \\
		\emph{RIS-MCTS + C (1s)} & 20.4 & 19.9 & 19.8 & 18.8 \\
		\emph{RIS-MCTS + C (10s)} & 20.6 & 19.8 & 19.7 & 18.5 \\
		\emph{RIS-MCTS + vdb (1s)}& 18.3 & 20.2 & 19.4 & 18.4 \\
		\emph{RIS-MCTS + vdb (10s)}& 20.0 & 21.0 & 20.2 & 19.3 \\
		\emph{RIS-MCTS + vdb + C (1s)} & 19.6 & 20.8 & 20.2 & 19.3 \\
		\emph{RIS-MCTS + vdb + C (10s)} & 20.5 & 22.0 & 21.3 & 20.0 \\
		\hline
		WTFWThat \cite{DeepMindHanabi} & 19.5 & \textbf{24.2} & \textbf{24.8} & \textbf{24.9} \\
		SmartBot \cite{DeepMindHanabi} & 23.0 & 23.1 & 22.2 & 20.3 \\
		\hline
	\end{tabular}
\end{table}

As shown in Table \ref{soa}, there is no significant difference between 1s and 10s of computation time for RIS-MCTS when using a random simulation policy. 
However, with a more informed Van Den Bergh simulation policy RIS-MCTS continues to improve with time budget and substantially bests the 1-ply Expectimax search \cite{Bouzy} that uses a similar time budget and playable convention.
Overall RIS-MCTS with the `playable now' convention yields what were, at the time of CIG 2018, state-of-the-art results in Hanabi. Two papers since then \cite{DeepMindHanabi, BAD} using Deep Reinforcement Learning provide a new benchmark and use two bots from outside the academic literature that beat RIS-MCTS. These are shown in the last two lines of Table \ref{soa}; \emph{WTFWThat} uses a hat-guessing variant that is legal under standard Hanabi rules, and SmartBot uses a number of hand-crafted conventions modelled on high-level human play.

\section{MCTS in 40 milliseconds}\label{Mirror}
For the CIG 2018 Hanabi competition, an agent must make each move within 40ms; in Table \ref{soa}, algorithms that would breach this limit are show in \emph{italics}. RIS-MCTS is therefore not going to work well if we need 3000ms per move (or even 100ms). To cater for this tight time budget we borrow ideas from Silver et al. \cite{AlphaZero} and Anthony et al. \cite{anthony2017} and run RIS-MCTS games offline, then use their output to train a classifier with standard supervised learning techniques; a similar approach was used for Monte Carlo search in Skat \cite{Skat}. We then use this trained classifier to make decisions in a competition game within the 40ms time limit.
	`
We try two variants of classifier, and in both cases run 500 games of RIS-MCTS with 30 seconds per move to gather the training data.
The first variant is a direct classifier to pick a rule. We record the rule(s) the algorithm uses at each game-state, along with a feature representation of that state. Each move in an MCTS training game generates a single input-output pair. A total of 11 base features, plus 4 features per player are used (`core' and `player' categories in Table \ref{features}). Note that multiple rules can be triggered, as the same move may be recommended by more than one rule. 

\begin{table}
		\caption{Features used to summarise each Hanabi game-state.}
	\label{features}
	\centering
	\begin{tabular}{|l|p{6cm}|}
		\hline
		Category & Name \& Description \\
		\hline
		Core & \emph{Score}: The current game score out of 25 \\
		 & \emph{Information}: Available information tokens (0 to 8)\\
		 & \emph{Lives}: Current lives (0 to 3)\\
		 & \emph{Deck}: The number of undrawn cards in the deck \\
	 	& \emph{MovesLeft}: The number of moves left if we have emptied the deck (if the deck is not empty, then this is the number of Players + 1) \\
		 & \emph{UnavailablePoints}: Points permanently lost due to discards \\
	 & \emph{FivesOnTable}: The number of 5s played successfully \\
	& \emph{FoursOnTable}:  The number of 4s played successfully \\
	& \emph{ThreesOnTable}:  The number of 3s played successfully \\
		& \emph{TwosOnTable}:  The number of 2s played successfully \\
		\hline
		Player & \emph{MaxPlayableProb}: The probability a card is playable (max over all cards) \\
		 & \emph{MaxDiscardableProb}: The probability a card is discardable (max over all cards)\\
		 & \emph{MaxPlayablePlusOneProb}: The highest probability that a card is almost playable (one more card of that colour must be played first) \\
		 & \emph{PlayerInfo}: The amount of positive information a player has (1 point per piece of information per card) \\
		\hline
		Action & \emph{Play}: 1 if the action is `Play Card', 0 otherwise \\
		 & \emph{PlayableProb}: probability that played card is playable (0 if not a Play action) \\
		 & \emph{PlayCompleteProb}: probability that played card will complete a colour (0 if not a Play action) \\
		 & \emph{Discard}: 1 is the action is `Discard Card', 0 otherwise \\
		 & \emph{DiscardableProb}: probability that discarded card is not one that can ever be played (0 if not a Discard) \\
		 & \emph{LastUsefulProb}: probability that discarded card is the last of a useful pair, or a 5 (0 if not a Discard) \\
		 & \emph{PointsForegone}: Expected number of points that will be unachievable when card is discarded (0 if not a Discard) \\
		\hline
	\end{tabular}
	\end{table}

This creates a set of input-output pairs (from features to the triggered rules), and we use a shallow neural network to learn a mapping from input to output. The network has 31 input neurons - one per feature, up to a 5-player game - and 9 output neurons, one per rule in Table \ref{RuleList}. There is a single hidden layer with 30 neurons.
The classifier is trained to minimise cosine similarity in the 9-dimensional output space.

The second variant learns a function approximator for the action-value function $Q(s,a)$, the value of taking action $a$ from state $s$. With this variant we record a feature representation that combines the current game state and action chosen, and the output we predict is the mean end-game score that we will obtain as a result. This adds 7 further `Action' features to Table \ref{features}. The network now has 38 input neurons, and a single output neuron to estimate the final game score that will result.
For Hint actions all the `Action' features are set to zero, and the Hint action is applied to the current game state to produce a roll-forward state. The `Core' and `Player' features are then extracted from this roll-forward state. This roll-forward is not possible (in general) for `Play' or `Discard' Actions, as the player does not know exactly what the card is - one possibility would be to calculate an expected roll-forward value by enumerating all possible states, or for a random sample of them, as in Expectimax search in \cite{Bouzy}. We adopt a less computationally demanding approach and extract features from only the current state, and add in the `Action' features. 
Each move in a training RIS-MCTS game will generate one input-output pair for each move that was considered from the root during tree search. The output target for each of these is set to be the mean score of the child node, i.e. the mean score over all MCTS iterations that took that action.
To use this function approximator as a classifier, we apply it separately to each of the possible actions implied by the rules in Table \ref{RuleList} and select the action that gives the highest predicted final score.

All Neural Networks were trained using DL4J \cite{DL4J}, with Rectified Linear activation at the hidden layer, and Rectified Tanh activations at the output. Squared error loss, Momentum of 0.9, a learning Rate of 1e-4 and L1 normalization of 1e-5 were found to provide the best results. The default Adam optimiser was used with a batch size of 16.
All inputs were normalized to a mean of zero and variance of one.

\begin{table}
	\caption{Hanabi results with 40ms limit, for 2, 3, 4 and 5 players. Bold entries show best results for each player count, and the standard error on all RIS-MCTS figures is $\pm 0.1$. See text for explanation of entries in lower section.}
	\label{40ms}
	\centering
	\begin{tabular}{|l|l|c|c|c|c|}
		\hline
		Algorithm & 2-P & 3-P & 4-P & 5-P \\
		\hline
		Ozawa et al. 2015  \cite{Ozawa} & 15.9 & - & - & - \\
		Van den Bergh 2015 \cite{vdb} & 13.8 & 17.7 & 17.2 & 16.3 \\
		Cox et al. 2015\cite{Cox}  & - & - & - & \textbf{24.7} \\
		Bouzy 2017 \cite{Bouzy} & 15.9 & 17.9 & 19.7 & 19.2 \\
		Canaan et al. 2018 \cite{Canaan}  & 20.1 & 19.6 & 19.4 & 18.3 \\
		Foerster et al. 2018 \cite{BAD} & \textbf{23.9} & - & - & -\\
		Bard et al. 2019 \cite{DeepMindHanabi} & 22.7 & 20.2 & 21.6 & 16.8 \\
		WTFWThat \cite{DeepMindHanabi} & 19.5 & \textbf{24.2} & \textbf{24.8} & \textbf{24.9}\\
		SmartBot \cite{DeepMindHanabi} & 23.0 & 23.1 & 22.2 & 20.3 \\
		\hline
		EvalFn (1) & 20.0 & 19.5 & 19.0 & 17.8 \\
		RIS-MTCS-NR (1) & 20.0 & 20.1 & 19.8 & 19.0 \\
		EvalFn (2) & 20.0 & 19.9 & 19.6 & 18.5 \\
		RIS-MTCS-NR (2) & 21.0 & 20.5 & 20.0 & 19.1 \\
		EvalFn (3) & 18.8 & 20.0 & 20.1 & 19.2 \\
		RIS-MTCS-NR (3) & 20.5 & 21.0 & 20.9 & 19.7 \\
		\hline
	\end{tabular}
\end{table}

Rather than just use the statistics at the root node at each stage of a Training Game, as in previous work \cite{AlphaZero} \cite{hearthstoneZero}, we also use statistics from deeper nodes in the tree. Each of these deeper nodes represents a game-state explored by the training game, and can be used to generate input-output training pairs in the same way as the root.
We only include nodes with at least 50 visits in the training data to avoid noise from nodes not fully explored.

A theoretical insight into why this is useful is that these deeper states form a penumbra of exploration away from the game actually played. If we restrict training to game-states that were actually encountered in the training games then we risk our classifier being fragile in play, as it will be sensitive to out-of-sample game states in which it may play badly (or just randomly). This is the insight behind, for example, the {\sc DAgger} algorithm in imitation learning \cite{Dagger}, and \cite{RossBagnell} show that this fragility from training on only the states encountered by the expert that we are learning to imitate leads in general to a final error that increases with $O(T^2)$ in the worst case, where $T$ is the number of sequential moves made. Correcting the training data to be the distribution that the learned classifier will meet in the real world reduces this dependence to $O(T)$ in the worst case, and our `penumbra of exploration' makes a step in this direction, allowing the trained classifier to function more intelligently in a wider range of game-states.

The resultant networks can be used to play Hanabi directly, triggering the valid rule at each stage that has the highest activation. This is much faster than the 40ms time limit, so to make productive use of this time we use the same approach as DeepMind's AlphaZero \cite{AlphaZero}. Within the time budget we conduct a brief MCTS with no simulations, and the value of new leaf nodes estimated using the learned value approximator $Q(s, a)$. We estimate $V(s)$ using $Q(s, a)$ by setting all the `Action' features to zero. This is the same as if we were valuing a Hint action that would lead to the current state after roll-forward. At a leaf node in the tree, all the child nodes are expanded at the same time instead of just one, as in\cite{progressiveWidening}, and the value of the highest estimate is used directly in back-propagation.  Swiechowski et al \cite{hearthstoneZero} use a similar approach in Hearthstone.



When a node is expanded, vanilla MCTS will record zero visits until back-propagation occurs through the node. Instead we can use domain knowledge to estimate the value of a node when it is expanded. This can be represented by a number of implied prior visits $n(a)$ and a mean score $V(a)$ for these visits to be used in later UCT selection using (\ref{UCT}).
A heuristic can be used to seed the value of initial $n(a)$ and $V(a)$ \cite{AlphaZero} \cite{GellySilver2007}. Like progressive un-pruning, this avoids spending time on exploring every action before we can start exploiting.
We initialise expanded nodes with $n(a) = 1$ and $V(a)$ equal to the estimate by the trained evaluation function.

Table \ref{40ms} summarises the results. `EvalFn (1)' is the value function trained on the results of 300 games using RIS-MCTS with random rollouts - all with 5 players, 3 seconds per move, and using all nodes with at least 50 visits, giving 258k records for supervised training. 
`EvalFn (2)' is then trained using 500 games using RIS-MCTS and `EvalFn (1)' as the simulation policy, with 10 seconds per move. Finally `EvalFn (3)' was trained using the output of 200 games, with 30 seconds per move, but with $C=3$ instead of $C=0.1$ used in the previous iterations. This change was designed to sample states from a broader penumbra of exploration around the base game trajectory, and generated 704k records for training. This broadening was necessary to obtain a performance improvement in the third iteration.
`RIS-MCTS-NR (n)' uses RIS-MCTS with 30ms per move\footnote[3]{We use 30ms per move rather than 40ms as contingency for processing overhead in other code and the competition framework.}, the `playable now' convention and the leaf node valued by `EvalFn (n)' with no further rollout (NR).
A similar interleaving of imitation learning and data generation is used in \cite{anthony2017}.

As Table \ref{40ms} shows, there is an increase in performance as we iterate on the evaluation function. In all cases a strategy using the trained evaluation function to define a greedy action played Hanabi at least as well as the RIS-MCTS algorithm that generated the training data, and often much better. For example in the first iteration, the training games (with 3 seconds per move) had a mean score of $14.8 \pm 0.3$, while `EvalFn (1)' scores $17.8 \pm 0.2$ with 5-players (with an average of 3ms per move). In the third iteration, RIS-MCTS with 30 seconds per move scored $19.3 \pm 0.1$ on 5-player games, the same as `EvalFn (3)' trained on the generated data.

Using even a 30ms budget for search significantly improves on the result, and gives the best results for Hanabi play on all variants other than 5-players, for which the Hat Guessing of \cite{Cox} is much better. Surprisingly,  if we compare with Table \ref{MCTSRuleResults} a 30ms budget here is as effective as 1000ms in full RIS-MCTS.


\section{Mixed Track}\label{Mixed}
In the mixed track we do not know what strategy the other players are using, and hence must model them to predict their future actions. The approach we take is based on Predictor IS-MCTS \cite{Walton-Rivers}, which uses IS-MCTS only for the moves of the current player, and predicts the moves of the other players with a heuristic policy. In this way, if we know what strategy they are using, we can model that as part of the environment and search for a good move given our prediction of their behaviour.

Given the 40ms time limit we use the approach from Section \ref{Mirror} and evaluate the leaf state using the trained EvalFn without simulations. Because the tree is not used to model the moves of other players, there is no requirement to determinize the game state from their perspective, and this is \emph{not} RIS-MCTS.

\begin{table}
	\caption{Policies used to model other player strategies in the Mixed Track. The first eight entries are taken from \cite{Walton-Rivers}}
	\label{allPolicies}
	\centering
	\begin{tabular}{|l|p{6cm}|}
		\hline
		Rule & Description \\
		\hline
		random & Completely random \\
		cautious & Only plays a card if 100\% certain it is playable \\
		iggi & A modification of `cautious' with a deterministic Discard rule \\
		flawed & Tells randomly, and will Play a card it is only slightly sure of \\
		piers & Based on `iggi', but with more sophisticated rules \\
		risky & Will play a card if only 60\% certain it is playable \\
		outer & The `outer' policy from \cite{Ozawa} \\
		Van Den Bergh & The best performing rule from \cite{vdb} \\
		\hline
		evalFn & A policy defined by the evaluation function trained in \ref{Mirror}. Each rule in Table \ref{RuleList} is evaluated, and the best one picked\\
		evalFn+C & As `evalFn', but using the convention that a single touch by the previous player means a card is playable\\
		\hline
	\end{tabular}
\end{table}
In the CIG 2018 competition we do not know the strategies of the other players, so we need to estimate them. We do this with a simple Bayesian model over the list of policies in Table \ref{allPolicies}. Eight of these are taken directly from \cite{Walton-Rivers}, with the addition of two policies defined by the evaluation functions trained in Section \ref{Mirror}. 

For each iteration of IS-MCTS one of the possible policies in Table \ref{allPolicies} is sampled for each player, using the current posterior distribution for that player. The prior for this distribution is initialized as a non-uniform categorical distribution with a log likelihood of -40 for all strategies except `EvalFn+C', which receives the remainder of the probability mass. This rather arbitrary bonus assumes better play than a uniform prior over all strategies, and was found to give better results, even when playing with a uniform distribution of other policies.

Once all the other players have made an actual move in the game, these are used to update the categorical distribution for each other player individually in a formal Bayesian update that assumes there are only 10 possible policies, and which learns a function to approximate the likelihood. The steps are:
\begin{enumerate}
	\item {Extract a feature representation (Table \ref{features}) of the move made and game state immediately prior from the perspective of the relevant player.}
	\item{Apply a learned shallow neural network to this to estimate the probabilities that the move was selected by each of the 10 strategies.}
	\item{Do a Bayesian update on the distribution by subtracting the estimated negative log-likelihood from each strategy parameter in the categorical distribution, and then re-normalizing.}
\end{enumerate}

To train the network used in step 2, 1000 games of Hanabi were run with random numbers of players (between 2 and 5), and with each player randomly assigned one of the 10 policies in Table \ref{allPolicies}. For each move a training tuple was created of the feature representation of the state and action taken, and the acting policy as a one-hot target. A network with one hidden layer of 30 units and a softmax over 10 output units was then trained using the same parameters as in Section \ref{Mirror}.

The log likelihood of the observed data arising from the policy is only an estimate.
To prevent a rogue estimate in step 2 causing the posterior log likelihood for a policy becoming too negative too quickly in step 3, this update was capped at $\log(10^{6})$. 

This approach to opponent modelling (here, actually ally modelling) is very similar to the Bayesian opponent model in a Reinforcement Learning environment in \cite{tesauroQL}.

\section{RIS-MCTS and Imperfect Information Games}\label{discussion}
As detailed in Section \ref{ISMCTSfailure}, as the number of iterations of MO-ISMCTS increases, it will model the actions of other players using perfect knowledge of the information known to the root player, albeit not of the whole game state. Information leaks from the root player to its opponent models and causes strategy fusion in the opponent's decisions in the tree.
RIS-MCTS re-determinizes at each node and avoids information known by the root player leaking into the opponent model within tree search.

This information leakage in IS-MCTS is particularly relevant to Hanabi because the unusual form of information hiding means we have knowledge about the other player's position that they do not. This means that the leakage strongly affects the next player's move directly, at depth two in the search tree. In the more general case where our hidden information is about our own position, this information leakage will be much weaker as it can only arise after our \emph{next} move much deeper in the tree.

The implicit assumption that other players know everything that we know means that there is no benefit to bluffing an opponent (e.g. to mislead them about what hand I have). It is for this reason that Cowling et al. \cite{Resistance}, introduce `fake' determinizations into opponent trees with the MT-ISMCTS algorithm. This better models the opponents' lack of full knowledge and incentivises bluffing behaviour. 
In the case of a co-operative game like Hanabi we wish to impart genuine information to the other players (`signalling' rather than `bluffing'), but the principle remains that there is no incentive to pass on information within MO-ISMCTS.
Online Counterfactual Regret methods \cite{Bitan} similarly sample game tree trajectories from the start of the game that do not lead to the current root game state. In all cases, sampling of game-states from outside the current game-subtree is essential to address non-locality: what might have happened earlier in the game, but didn't, affects the optimal decision now given that imperfect information extensive games are not sub-game perfect. 
RIS-MCTS does not address the non-locality problem in any new way.

By re-determinizing at each node, RIS-MCTS can create determinized game states that are incompatible with the root information set, and introduces `fake' determinizations as in MT-ISMCTS. The differences are that we do not need to maintain one opponent IS-MCTS tree for each of their possible information sets (which would be infeasible in Hanabi) and that these incompatible determinizations are created during each MCTS iterations as mutations from the root determinization. 
The introduction of non-compatible determinizations at opponent nodes in RIS-MCTS is needed to avoid strategy fusion, but leads to the back-propagation of impossible game results to the root node. As Cowling et al. note in \cite{Resistance}, use of invalid determinizations in this way pollutes the tree statistics, and may lead to poorer decisions. We speculate this may contribute to the slight decline in performance seen for RIS-MCTS for larger computational budgets in Table \ref{soa}.
This is a significant theoretical weakness in RIS-MCTS as described here, since the result of a determinized game should in principle only be back-propagated to nodes which that game could actually have originated. 
We intend to address this aspect of RIS-MCTS, and investigate its performance in games with more classic forms hidden information such as the phantom games also used as a test-bed in \cite{IS-MCTS}.

\section{Conclusion} \label{conclusion}

We have analysed why vanilla IS-MCTS performs very poorly in Hanabi, and presented in RIS-MCTS an extension that re-determinizes at each node from the perspective of the active player at the node. RIS-MCTS thus avoids the information known by the root player leaking into the opponent model within tree search and the strategy fusion that arises from this. With RIS-MCTS, and a restriction of the action space using rule heuristics we achieve the best results so far in Hanabi (apart from in a 5-player game, where the `hat-guessing' of \cite{Cox} is much better). 
Nevertheless we have highlighted in Section \ref{discussion} a theoretical flaw in RIS-MCTS, especially compared to MT-MCTS \cite{Resistance}, and future work is intended to address this. 

To improve play in Hanabi specifically a number of avenues present themselves:
\begin{itemize}
	\item {investigate the power of conventions other than the single one used here;}
	\item {a convention is a hard form of inference on possible game-states, and RIS-MCTS still assumes a uniform distribution of game states within the root information set. Further work could look at softer inference methods to maintain a more sophisticated set of beliefs about cards in hand;}
	\item{investigate how much iteration and broadening of exploration is helpful when using the statistics of a MCTS tree to train an evaluation function.}
	\item {the use of heuristic rules to restrict the action space could naturally be extended with progressive un-pruning \cite{progressiveWidening}, using the learned evaluation function to decide which actions to un-prune.}
\end{itemize}

\section{Acknowledgments}
I would like to thank Joseph Walton-Rivers and Piers Williams for organising the CIG 2018 Hanabi competition. They and Rodrigo Canaan provided several fruitful discussions around Hanabi.

\bibliography{Hanabi}
\bibliographystyle{ieeetran}

\end{document}